\documentclass[runningheads]{llncs}
\usepackage[section]{placeins}
\usepackage[T1]{fontenc}
\usepackage{tcolorbox}
\usepackage{listings}
\usepackage{enumitem}
\usepackage{xcolor}
\usepackage{graphicx}
\usepackage{booktabs}
\usepackage{multirow}
\usepackage[misc]{ifsym}
\newcommand{\corr}{(\Letter)}
\usepackage{amsmath}
\usepackage{amssymb}
\usepackage{algorithm}
\usepackage{algpseudocode}
\usepackage{url}
\usepackage[hidelinks]{hyperref}

\newcommand{\pp}{\,pp}

\begin{document}

\title{ASDA: Automated Skill Distillation and Adaptation\\for Financial Reasoning}

\titlerunning{ASDA: Automated Skill Distillation and Adaptation for Financial Reasoning}

\author{Tik Yu Yim\corr \and Wenting Tan \and Sum Yee Chan \and
        Tak-Wah Lam \and Siu Ming Yiu}
\authorrunning{T. Y. Yim et al.}
\institute{The University of Hong Kong, Hong Kong SAR, China \\
\email{\{tyyim, wt212796\}@connect.hku.hk, sumychan@hku.hk} \\
\email{\{twlam, smyiu\}@cs.hku.hk}}
\toctitle{ASDA: Automated Skill Distillation and Adaptation for Financial Reasoning}
\tocauthor{Tik Yu Yim, Wenting Tan, Sum Yee Chan, Tak-Wah Lam, Siu Ming Yiu}
\maketitle

%----------------------------------------------------------------------
% ABSTRACT
%----------------------------------------------------------------------

\begin{abstract}

Adapting large language models (LLMs) to specialized financial reasoning typically requires expensive fine-tuning that produces model-locked expertise. Training-free alternatives have emerged, yet our experiments show that leading methods (GEPA and ACE) achieve only marginal gains on the FAMMA financial reasoning benchmark, exposing the limits of unstructured text optimization for complex, multi-step domain reasoning. We introduce Automated Skill Distillation and Adaptation (ASDA), a framework that automatically generates structured skill artifacts through iterative error-corrective learning without modifying model weights. A teacher model analyzes a student model's failures on financial reasoning tasks, clusters errors by subfield and error type, and synthesizes skill files containing reasoning procedures, code templates, and worked examples, which are dynamically injected during inference. Evaluated on FAMMA, ASDA achieves up to +17.33\pp{} improvement on arithmetic reasoning and +5.95\pp{} on non-arithmetic reasoning, substantially outperforming all training-free baselines. The resulting skill artifacts are human-readable, version-controlled, and compatible with the Agent Skills open standard, offering any organization with a labeled domain dataset a practical and auditable path to domain adaptation without weight access or retraining\footnote{Code and skill libraries are available at \url{https://github.com/SallyTan13/ASDA-skill}}.

\end{abstract}

%======================================================================
\section{Introduction}
\label{sec:intro}
%======================================================================

Financial reasoning poses a distinctive challenge for general-purpose LLMs: it demands simultaneous mastery of multi-step quantitative calculation and deep domain-specific judgment---a combination that pure math or pure knowledge benchmarks do not jointly test~\cite{xue2024famma,xie2024finben}. Evaluations across multiple financial benchmarks confirm a persistent performance ceiling: FAMMA reveals that standard, non-reasoning frontier models achieve only 38--45\% overall accuracy across eight financial subfields~\cite{xue2024famma}\footnote{Extended-thinking models (GPT-o1, DeepSeek-R1, Qwen-QwQ-32B) score 67–76\% on FAMMA, and PoT-augmented variants reach 78–86\%. We exclude these as thinking budget introduces a confound orthogonal to domain adaptation; our evaluation targets standard non-reasoning models under fixed inference budgets.}; and FinBen finds that while LLMs handle information extraction well, they consistently struggle with advanced reasoning and complex financial QA~\cite{xie2024finben}. FAMMA's error analysis finds that \emph{domain-knowledge gaps} 
dominate model errors~\cite{xue2024famma}---models misapply financial concepts to the wrong context or lack the expertise to select the correct procedure. These are not failures that more parameters alone will fix---they require targeted advances in domain reasoning. The standard remedy---domain-specific fine-tuning---is costly, produces \emph{model-locked expertise} that becomes obsolete with each model release, and depends on supervision resources that many organizations lack~\cite{wu2023bloomberggpt,yang2023fingpt}. This is especially problematic in regulated industries such as financial services, legal, and healthcare, where organizations deploy commercial LLMs via black-box API without weight access.  Automated prompt optimization offers a training-free alternative, but methods like GEPA~\cite{agrawal2025gepa} and ACE~\cite{ace2025} optimize \emph{flat text strings} that cannot capture the structured, multi-step procedural knowledge financial reasoning demands---a limitation our FAMMA experiments confirm, where both methods achieve only marginal gains (Section~\ref{sec:results}). The missing abstraction is not a better prompt, but an \emph{executable skill}: a modular, self-contained reasoning procedure that can be independently composed, tested, and updated for each target domain.

We introduce \textbf{Automated Skill Distillation and Adaptation (ASDA)}, a framework that automatically generates executable agent skills from error analysis without modifying model weights. A teacher model diagnoses a student's failures on financial tasks and clusters them by subfield and error type to identify the root causes of \emph{domain-knowledge gaps}. From these clusters, it synthesizes skill files containing domain-specific reasoning procedures and code templates, which a selector injects at inference time. Evaluated on FAMMA, ASDA achieves up to \textbf{+17.33\pp{} on arithmetic} and \textbf{+5.95\pp{} on non-arithmetic reasoning} after iterative refinement, significantly outperforming all training-free baselines. The resulting skill library is not a better prompt---it is a new representational layer between the model and its deployment context that can be version-controlled, audited, and regenerated for any successor model.

Our contributions are as follows:

\textbf{(1) ASDA framework.} We introduce the first system to automatically generate executable agent skills for domain-specific reasoning using only black-box LLM access---no weight updates or gradient 
computation---substantially outperforming all training-free baselines on FAMMA.

\textbf{(2) Self-sufficient adaptation from questions and answers alone.} A self-teaching ablation shows that ASDA can improve a model using only the questions and ground-truth answers in the training set---no superior teacher model required---achieving +6.33\pp{} (73\% of the full gain). This means any organization with a labeled domain dataset can run ASDA on their deployed model directly, without access to a stronger or more expensive model, making the framework practical for real-world enterprise deployment.

%======================================================================
\section{Related Work}
\label{sec:related}
%======================================================================

\subsection{Financial LLM Adaptation}

Domain-specific fine-tuning has been the dominant approach to adapting LLMs for financial tasks. BloombergGPT~\cite{wu2023bloomberggpt} demonstrated the potential of finance-specific pretraining but required approximately 1.3 million GPU hours on a proprietary 363B-token corpus---resources beyond most organizations. FinGPT~\cite{yang2023fingpt} offered a more accessible alternative through LoRA-based fine-tuning on open financial data. More recently, Xue et al.~\cite{xue2024famma} explored distillation from DeepSeek-R1 to smaller models for financial reasoning. Despite these advances, all fine-tuning approaches share a fundamental limitation: they produce model-locked expertise that requires re-training when the base model is updated or replaced, and many are incompatible with black-box API access to commercial LLMs.

\subsection{Training-Free Adaptation}

Automated prompt optimization offers a weight-free alternative. Prior work formalizes this as automatic differentiation over text~\cite{yuksekgonul2024textgrad} or optimizable module compositions~\cite{khattab2023dspy}. GEPA~\cite{agrawal2025gepa} achieves state-of-the-art results on several benchmarks through reflective prompt evolution. ACE~\cite{ace2025} takes a test-time knowledge accumulation approach, building contextual expertise during inference. While these methods avoid fine-tuning costs, their output remains a flat text string---a monolithic instruction block applied uniformly to all questions---that cannot represent distinct reasoning procedures for different financial subdomains or error types.

A complementary line of work treats failure analysis as the primary learning signal. LEMMA~\cite{ding2025lemma} synthesizes error-type-grounded training data for mathematical reasoning and consistently outperforms correction-agnostic data augmentation baselines. This establishes that failure-driven analysis yields richer adaptation signal---a principle ASDA extends to training-free, executable skill generation. At the other end of the adaptation spectrum, test-time training (TTT) methods temporarily update model weights at inference, and class incremental approaches continually update them during deployment~\cite{Chen2026add}, requiring gradient access incompatible with black-box API deployments; ASDA occupies the gap between these approaches.

\subsection{Skill-Based Agent Architectures}

Recent agent work has established skills as reusable, first-class knowledge
artifacts. Voyager~\cite{wang2023voyager} demonstrated that composable
executable skills---stored as JavaScript code---enable continual capability
growth in open-ended Minecraft environments without parameter updates. The
Agent Skills open standard~\cite{anthropic2025skills} formalized portable
Markdown skill files with routing metadata, progressive disclosure, and
embedded code templates; ASDA generates skill artifacts compatible with this
standard.

Recent benchmarking work quantifies when skills help or hurt. SkillsBench~\cite{li2026skillsbench} finds that curated skills raise average pass rate by $+16.2\pp{}$ across 11 domains, while self-generated skills provide no benefit on average---models cannot reliably author the procedural knowledge they benefit from consuming. SWE-Skills-Bench~\cite{han2026sweskills} reports that 39 of 49 public SWE skills produce no measurable improvement (average gain: $+1.2\%$), with three actively degrading performance due to context-mismatched guidance. Together, these results establish that skill--task alignment is a precondition for benefit---the gap ASDA's error-driven distillation addresses by generating skills calibrated to a specific model's failure distribution on a target domain.

Several concurrent systems also explore automated skill generation from failure signals. EvoSkill~\cite{alzubi2026evoskill} iteratively diagnoses execution failures in coding agents, using ground-truth-supervised feedback in an online interactive loop to propose and validate new skill folders. Trace2Skill~\cite{ni2026trace2skill} collects execution logs across many tasks and analyzes them in parallel to extract recurring patterns into reusable skills, reporting strong cross-model and out-of-distribution transfer. SkillRL~\cite{xia2026skillrl} co-evolves skills with a reinforcement learning policy, requiring SFT and weight updates. Each of these systems depends on execution traces, trajectory logs, or gradient access. ASDA requires none of these: it distills skills from static question--answer pairs alone, introduces a dual-phase refinement that explicitly balances coverage gains against regression risk, and operates entirely through black-box API access. To our knowledge, no prior work generates executable skills for domain-specific reasoning in a fully training-free setting from static question--answer pairs alone.

%======================================================================
\section{Method: ASDA Framework}
\label{sec:method}
%======================================================================

\begin{figure}[!t]
\includegraphics[width=1.02\textwidth, trim=5 515 5 25, clip]{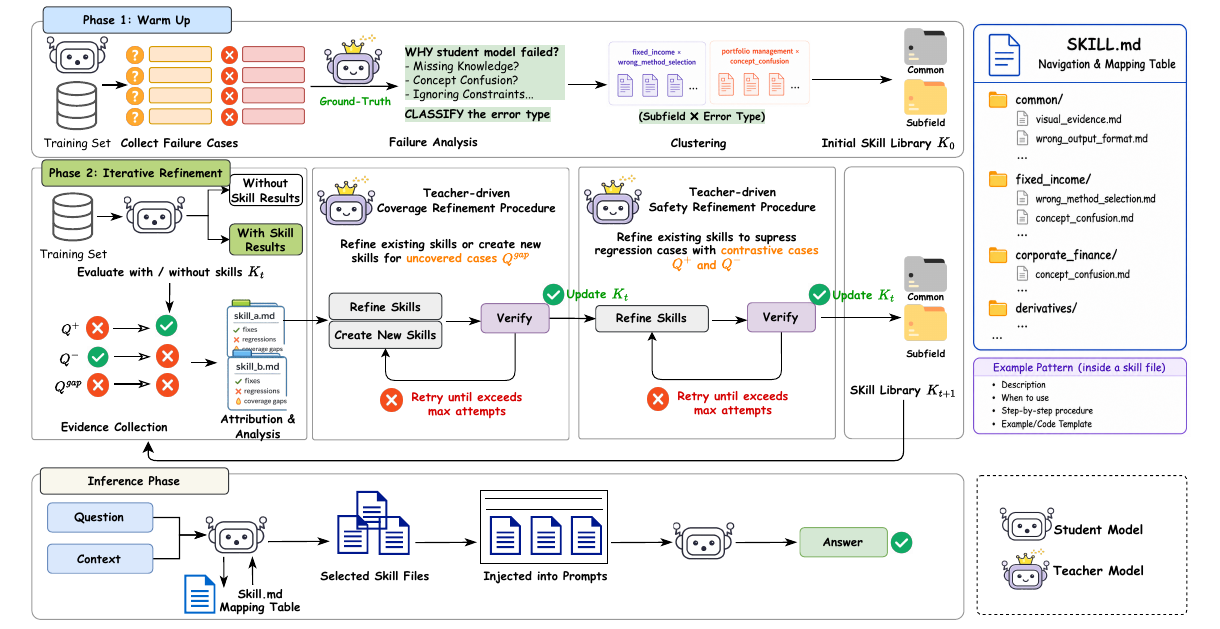}
\caption{Overview of the ASDA framework. \textbf{Phase~1 (warm-up):} the teacher model analyzes student failures, producing structured annotations that are clustered by subfield and error type to synthesize an initial skill library $\mathcal{K}_0$. \textbf{Phase~2 (iterative refinement):} the library is refined through three sequential procedures, evidence collection and attribution procedure, coverage refinement procedure (resolving uncovered failures in $Q^\mathrm{gap}$) followed by safety refinement procedure (suppressing regressions in $Q^-$), with every skill update gated by a correctness threshold. \textbf{Inference Phase:} a selector reads \texttt{SKILL.md} and injects the relevant skill files into the student's prompt.}
\label{fig:pipeline}
\end{figure}

ASDA follows a teacher--student architecture with two main phases: (1)~a \emph{warm-up} that establishes an initial skill library from systematic failure analysis, and (2)~an \emph{iterative refinement} loop that automatically refines the skill library through repeated evidence collection, skill revision, and validation. Figure~\ref{fig:pipeline} provides an overview of the full framework.

Before describing each phase, we introduce three terms used throughout the paper. An \textbf{error type} is one of ten categories from a predefined
taxonomy.\footnote{Error types are: visual evidence, wrong method
selection, concept confusion, missed multi-step computation, unit/currency
mistakes, missed constraints, wrong targets, wrong output format, code
execution errors (PoT-specific), and other. The taxonomy was generated by prompting an LLM to categorize
recurring failure patterns from the student's errors on the training set,
then refined through author review.} A \textbf{pattern} is a single, named failure scenario within a skill file, one specific knowledge gap or recurring mistake the model makes. A \textbf{skill file} groups all patterns that share the same financial subfield and error type; for example, \path{fixed_income/wrong_method_selection.md} collects all wrong-method failures observed in the fixed income subfield.

%----------------------------------------------------------------------
\subsection{Warm-Up}
%----------------------------------------------------------------------

\subsubsection{Failure Analysis and Structured Annotation}

The warm-up phase constructs an initial skill library $\mathcal{K}_0$ from the student model's failures on the training set. For each question the student answers incorrectly, the teacher model receives the question, the student's incorrect answer and reasoning trace, and the ground-truth answer. The teacher is then prompted to perform failure analysis and output a structured annotation in the following format:

\begin{lstlisting}[
  basicstyle=\ttfamily\small,
  frame=single,
  backgroundcolor=\color{gray!8},
  columns=flexible,
  breaklines=true,
  aboveskip=4pt, belowskip=4pt
]
{
  "subfield":    "fixed_income",
  "error_type":  "wrong method selection",
  "root_cause":  "Lacks knowledge that forward rates must be composed 
  sequentially as discount factors, not applied independently per period"
}
\end{lstlisting}

% \noindent The \texttt{error\_type} field is constrained to the ten-type taxonomy defined above. The \texttt{root\_cause} field is the most consequential: the teacher is explicitly prompted to identify the underlying knowledge gap, the specific concept the model is missing---rather than a surface description of what was computed incorrectly. This forces the diagnosis to be actionable for skill synthesis. The \texttt{required\_ops} field records the operations needed to solve the question correctly, supporting downstream skill routing.
The \texttt{error\_type} field is constrained to the taxonomy defined above. The \texttt{root\_cause} field captures the underlying knowledge gap rather than a surface description of what was computed incorrectly, ensuring the diagnosis is actionable for skill synthesis.

\subsubsection{Skill Library Organization}

The annotated failures are clustered by their \texttt{(subfield, error\_type)} pair. Each cluster becomes one skill file. The library is therefore organized as a two-level hierarchy:

\begin{lstlisting}[
  basicstyle=\ttfamily\small,
  frame=single,
  backgroundcolor=\color{gray!8},
  columns=flexible,
  breaklines=false,
  aboveskip=4pt, belowskip=4pt
]
SKILL.md                           % navigation + mapping table
common/
  visual_evidence.md               % cross-subfield patterns
  wrong_output_format.md
fixed_income/
  wrong_method_selection.md        % one file per subfield x error type
  concept_confusion.md
corporate_finance/
  concept_confusion.md
derivatives/ 
  ...
\end{lstlisting}

\begin{figure*}[ht!]
\centering
\includegraphics[width=1\textwidth]{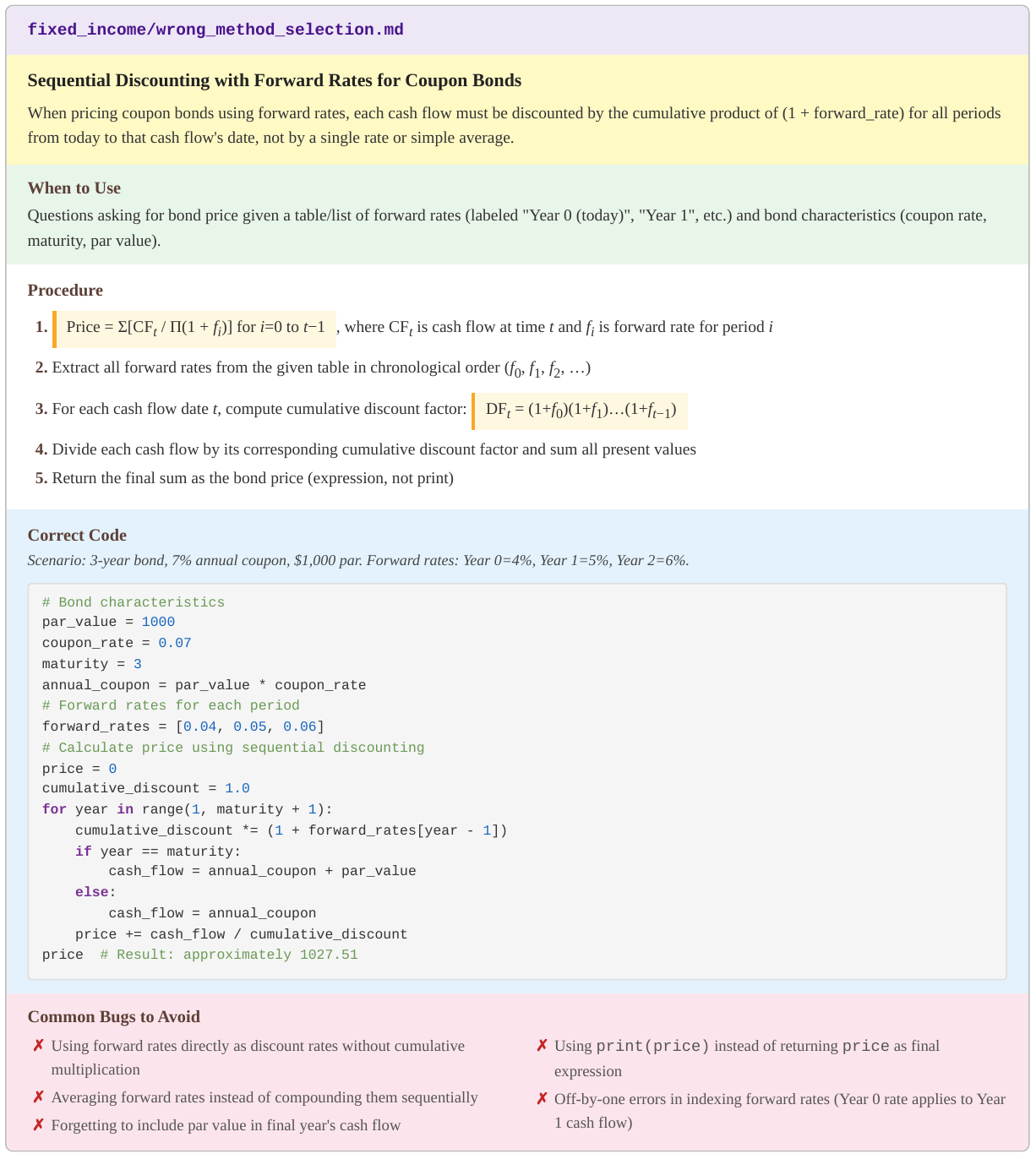}
\caption{One pattern from an ASDA skill file produced during warm-up from Haiku~3.5 failure analysis by Sonnet~4.5. The full file contains five additional patterns.}
\label{fig:skill-example}
\end{figure*}

Within each skill file, the teacher synthesizes one \emph{pattern} per distinct failure scenario identified in the cluster. Each pattern contains: a concise description of the addressed knowledge gap, explicit ``when to use'' conditions, step-by-step reasoning procedures, and worked examples or code templates. Figure~\ref{fig:skill-example} shows one pattern from \path{fixed_income/wrong_method_selection.md}; the full file contains five additional patterns covering other recurring failure scenarios in that subfield.

The library includes a top-level \texttt{SKILL.md} navigation file that summarizes the scope of each skill and provides a structured mapping from subfield keywords and failed financial patterns to skill file paths. The navigation file allows the downstream selector to identify relevant skills without parsing every skill file in full.

\subsubsection{Skill Selection and Injection}

At inference time, an LLM-based selector\footnote{In our experiments, we instantiate the selector using the same model as the student model. The selector is modular, and can be replaced by other LLMs.} reads the question text and the \texttt{SKILL.md} mapping table to identify the relevant financial subfield and match the question's characteristics against the listed patterns. The selector may choose multiple skill files for a single question, since complex financial reasoning often requires combining guidance from several patterns (e.g., a fixed income pricing question may need both \path{wrong_method_selection.md} and \path{common/missed_constraints.md}). Selected skill files are injected into the student's prompt as domain knowledge, guiding it to follow the appropriate reasoning procedure for the question.

%----------------------------------------------------------------------
\subsection{Iterative Refinement}
\label{sec:refinement}
%----------------------------------------------------------------------

The warm-up phase captures the most systematic failure patterns but leaves two residual problems. The first is \emph{coverage gaps}: some failures remain because existing skills do not yet address the required knowledge. The second is \emph{regressions}: skill injection causes previously correct answers to become incorrect when a skill overfits a narrow failure pattern and misleads the model on related but distinct questions. To address these issues, each refinement iteration begins with an \emph{evidence collection and attribution procedure}, which attributes skill-induced fixes, regressions, and unresolved gaps to individual skill files. The attributed evidence then drives two refinement procedures: a \emph{coverage refinement procedure} for expanding skill coverage and a \emph{safety refinement procedure} for suppressing regressions. In both procedures, every candidate skill update must pass through a \emph{verification gate}. The student re-solves the target questions with the updated skill injected, and the update is committed only if accuracy meets a predefined threshold~$\tau$. If it fails, the teacher regenerates a revised proposal, repeating up to $N_{\max}$ attempts before falling back to the previous skill version. The full process is summarized in Algorithm~\ref{alg:refinement}.

\subsubsection{Evidence Collection and Attribution}

At the start of each refinement iteration $t$, every training question is evaluated under two conditions: once with the current skill library $\mathcal{K}_t$ injected, and once without any skills. We compare the paired outcomes to derive three disjoint groups: $Q_t^{+}$ (correct with skills but incorrect without skills), $Q_t^{-}$ (incorrect with skills but correct without skills, i.e., regressions introduced by the current library), and $Q_t^{\mathrm{gap}}$ (incorrect under both conditions, i.e., coverage failures that the library has not yet resolved).

Since multiple skill files may be loaded for each question, a na\"{i}ve per-question outcome cannot be attributed to a specific file. An \emph{attribution} step follows: for each question in $Q_t^{+}$, $Q_t^{-}$, and $Q_t^{\mathrm{gap}}$, the teacher examines the student's reasoning trace alongside the loaded files and identifies the single file most responsible for the outcome. Questions are then re-grouped by their attributed file, producing per-file evidence sets that isolate each file's individual contribution to fixes, regressions, and remaining coverage gaps.

\begin{algorithm}[!t]
\caption{Iterative Refinement}\label{alg:refinement}
\begin{algorithmic}[1]
\Require Skill library $\mathcal{K}_0$, training set $\mathcal{D}$, student $M_s$, teacher $M_t$, max iterations $T$
\Ensure Refined skill library $\mathcal{K}^*$
\For{$t = 1$ to $T$}
    \Statex \hspace{\algorithmicindent}\textcolor{gray}{\textit{// Evidence Collection and Attribution}}
    \State $Q^+, Q^-, Q^{\mathrm{gap}} \gets \Call{EvalWithWithout}{\mathcal{D}, \mathcal{K}_t, M_s}$
    \State $\Call{AttributeToFiles}{Q^+, Q^-, Q^{\mathrm{gap}}, M_t}$
    \Statex
    \Statex \hspace{\algorithmicindent}\textcolor{gray}{\textit{// Coverage Refinement Procedure}}
    \For{each file $f$ with attributed $Q^{\mathrm{gap}}_f \neq \emptyset$}
        \State $f' \gets M_t.\Call{ProposeExpansion}{f,\; Q^{\mathrm{gap}}_f}$
        \If{$\Call{Verify}{f', Q^{\mathrm{gap}}_f, M_s} \ge \tau_{\mathrm{cov}}$}
            \State $\mathcal{K}_t[f] \gets f'$
        \EndIf
    \EndFor
    \State $\tilde{Q}^+, \tilde{Q}^- \gets \Call{PostCoverageVerify}{\mathcal{K}_t, Q^+, Q^{\mathrm{gap}}, M_s}$
    \Statex
    \Statex \hspace{\algorithmicindent}\textcolor{gray}{\textit{// Safety Refinement Procedure}}
    \For{each file $f$ with attributed $\tilde{Q}^-_f \neq \emptyset$}
        \State $f' \gets M_t.\Call{ProposeRepair}{f,\; \tilde{Q}^+_f,\; \tilde{Q}^-_f}$
        \If{$\Call{Verify}{f', \tilde{Q}^+_f, M_s} \ge \tau_{\mathrm{safe}}$}
            \State $\mathcal{K}_t[f] \gets f'$
        \EndIf
    \EndFor
    \State $\mathcal{K}_{t+1} \gets \mathcal{K}_t$
\EndFor
\State \Return $\mathcal{K}_T$
\end{algorithmic}
\end{algorithm}

\subsubsection{Coverage Refinement Procedure}

For each skill file, the teacher examines the $Q_t^{\mathrm{gap}}$ cases attributed to that file and diagnoses why coverage fails. Common causes include missing procedures for an edge case, trigger conditions that are too narrow to fire on relevant questions, or the absence of a worked example for a pattern the file does not yet contain. Based on this diagnosis, the teacher proposes an update, either by refining an existing pattern or by adding a new one. Each proposal is submitted to the verification gate: the student re-solves the attributed $Q_t^{\mathrm{gap}}$ cases with the candidate update injected, and the update is accepted only if the recovery rate exceeds $\tau_{\mathrm{cov}}$.

After all per-file coverage updates are committed, a post-coverage verification pass re-evaluates the entire affected set. Cases in $Q_t^{\mathrm{gap}}$ that are now solved are promoted into an updated positive set $\tilde{Q}_t^{+}$. Cases in $Q_t^{+}$ that regress under the modified skills are merged with the existing $Q_t^{-}$ to form an updated regression set $\tilde{Q}_t^{-}$. This updated partition forms the input to the safety refinement procedure.

\subsubsection{Safety Refinement Procedure}

The safety refinement procedure resolves regressions in $\tilde{Q}_t^{-}$, including both pre-existing regressions and those newly introduced by the coverage refinement procedure, while preserving correct behavior on $\tilde{Q}_t^{+}$.
For each skill file, the teacher receives both sets as contrastive evidence: the $\tilde{Q}_t^{+}$ cases, annotated with what the current skill gets right, serve as preservation constraints, whereas the $\tilde{Q}_t^{-}$ cases, annotated with what goes wrong, serve as repair targets. The teacher proposes a revised skill that removes or narrows the guidance responsible for the regressions while preserving the reasoning steps that produce correct answers on the positive set. Each proposal again passes through the verification gate with threshold $\tau_{\mathrm{safe}}$, which requires that accuracy on the positive cases remain above the allowed preservation threshold while improving performance on the negative cases.

After both refinement procedures complete, the updated library $\mathcal{K}_{t+1}$ becomes the input for the next iteration. 
% In practice, we find that performance peaks at epoch~2 before regressing at epoch~3 (see Section~\ref{sec:main_results}), indicating that two refinement passes represent the practical optimum.

%======================================================================
\section{Experimental Setup}
\label{sec:setup}
%======================================================================

\subsection{Benchmark: FAMMA}

FAMMA-Basic~\cite{xue2024famma} provides the scale, arithmetic/non-arithmetic decomposition, and self-contained textual context our pipeline requires.\footnote{We also considered FinMR~\cite{deng2025finmr}, FinanceMath~\cite{zhao2024financemath}, FinanceQA~\cite{mateega2025financeqa}, and FinMME~\cite{luo2025finmme}, but these are either too small for reliable train--test splits, withhold ground-truth answers for their primary test sets, yield baseline results that differ substantially from published figures, or focus on visual and retrieval-based reasoning rather than procedural financial reasoning.} It comprises 1,945 questions sourced from university textbooks and professional finance exams, spanning eight financial subdomains (e.g., corporate finance, derivatives, portfolio management) across three difficulty levels. An explicit decomposition between arithmetic and non-arithmetic questions enables separate evaluation of procedural and conceptual skill effectiveness. We use the FAMMA-Basic-Txt release, which provides OCR-extracted textual context for each question, ensuring that evaluation targets reasoning ability rather than retrieval.\footnote{FAMMA also includes a LivePro subset of 103 expert-curated questions, but only 35 are English-language and most are open-ended, making it too small for our pipeline.}

\subsection{Data Filtering and Split}

We restrict the corpus to the 1,378 English-language questions to control for language variation, ensuring that observed performance differences reflect reasoning capability alone. Since FAMMA provides no official train--test split, we construct our own: separating the English corpus into arithmetic and non-arithmetic subsets and applying stratified 60/40 splits based on difficulty level (easy, medium, hard) and question type (multiple-choice vs.\ open-ended). This produces 448 training and 300 test questions for arithmetic, and 378 training and 252 test questions for non-arithmetic. All reported results are on these held-out test sets and are therefore not directly comparable to those reported by Xue et al.~\cite{xue2024famma}.

\subsection{Evaluation Protocol}

ASDA's distillation pipeline operates on individual question--answer pairs, so we evaluate each question independently rather than in grouped LLM calls as in the original FAMMA protocol. FAMMA stores shared context only in the first sub-question of each group, so we propagate this context to all sub-questions to preserve information completeness.\footnote{Additional preprocessing: we enforce expression-based PoT 
outputs rather than \texttt{print()} statements to prevent execution 
failures.} Arithmetic questions use Program-of-Thought (PoT) code execution, with an additional selection step that maps numeric outputs to the closest multiple-choice option where applicable. For evaluation, we adopt a hybrid approach: rule-based exact matching for multiple-choice questions and LLM-based judging for open-ended questions, replacing the original protocol's use of LLM judging for all question types. Table~\ref{tab:eval-protocol} summarizes these modifications.

\begin{table}[!t]
\centering
\caption{Evaluation protocol modifications.}
\label{tab:eval-protocol}
\small
\begin{tabular}{lll}
\toprule
& Original FAMMA & Ours \\
\midrule
Question handling & Grouped sub-questions & Independent \\
MC evaluation & LLM judge & Rule-based exact match \\
Open-ended evaluation & LLM judge (GPT-4o) & LLM judge (Qwen-Max) \\
Arithmetic execution & PoT & PoT + MC mapping step \\
\bottomrule
\end{tabular}
\end{table}

We adopt Qwen-Max rather than GPT-4o as the open-ended judge for
cost\footnote{Qwen-Max via DashScope: \$1.60\,/\,\$6.40 per M tokens
in/out; GPT-4o: \$2.50\,/\,\$10.00 per M tokens in/out.}. Validation confirms
these modifications do not inflate gains: swapping Qwen-Max for
GPT-4o yields 99.3\% agreement across 1,104 judgments (delta:
0.00\pp{} arithmetic, $-0.39$\pp{} non-arithmetic); adopting the
full FAMMA LLM-judge strategy confirms the same shift.

\subsection{Baselines}

We compare ASDA against two leading training-free adaptation methods that operate under the same deployment constraint---black-box API access without weight modifications: GEPA~\cite{agrawal2025gepa} and ACE~\cite{ace2025}. Both methods optimize a single monolithic text applied uniformly to all questions. The baseline condition uses the student model with a standard task prompt and no injected skills or optimized prompts. Neither GEPA nor ACE has previously been evaluated on FAMMA; the results reported here are, to our knowledge, the first published evaluations of both methods on this benchmark.

To ensure a fair comparison, we adapt training conditions to each method's architectural requirements.\footnote{GEPA follows its original 3-way split protocol, training on 50\% of our training pool (222 arithmetic, 188 non-arithmetic) and selecting the best prompt on the remaining 50\%. ACE and ASDA use the full training pool (448 arithmetic, 378 non-arithmetic).}

\subsection{Implementation Details}

Two components are fixed across all experimental conditions: Qwen-Max serves as the LLM evaluation judge for open-ended questions, and Qwen-Turbo performs the MC selection step that maps numeric PoT outputs to answer choices. All inference is run at temperature 0 for reproducibility.\footnote{All models are accessed via the Anthropic API, with two exceptions: Haiku~3.5 via OpenRouter (no longer available on the Anthropic API directly) and Qwen models via DashScope.} 

Table~\ref{tab:cost} reports token usage and estimated cost for both the
warm-up pipeline and iterative refinement through two epochs (the paper's
best operating point). Warm-up cost is dominated by skill-augmented
evaluation; refinement is substantially more expensive due to repeated
train-and-test evaluation and teacher-driven analysis at each epoch.

\begin{table}[t]
\centering
\caption{ASDA pipeline cost (Haiku~3.5 student, Sonnet~4.5 teacher).
Teacher priced at Sonnet~4.5 rates (\$3\,/\,\$15 per M tokens in/out);
student at Haiku~3.5 rates (\$0.80\,/\,\$4 per M tokens in/out).
Qwen-based judging and MC selection tokens are included in student-phase
totals and conservatively priced at Haiku~3.5 rates; actual DashScope
costs are approximately 10$\times$ lower. Rates as of March~2026.}
\label{tab:cost}
\small
\setlength{\tabcolsep}{4pt}
\begin{tabular}{ll rr rr r}
\toprule
& & \multicolumn{2}{c}{\textbf{Teacher (S4.5)}}
& \multicolumn{2}{c}{\textbf{Student (H3.5)}}
& \\
\textbf{Config} & \textbf{Stage}
& \textbf{In} & \textbf{Out}
& \textbf{In} & \textbf{Out} & \textbf{Cost} \\
\midrule
\multirow{2}{*}{Arith}
  & Warm-up          & 545K  & 121K  & 8.07M  & 331K  & \$11 \\
  & Refine (E1--E2)  & 16.1M & 2.1M  & 65.4M  & 741K  & \$135 \\
\midrule
\multirow{2}{*}{Non-Arith}
  & Warm-up          & 485K  & 94K   & 5.80M  & 194K  & \$8 \\
  & Refine (E1--E2)  & 9.2M  & 1.2M  & 51.2M  & 598K  & \$89 \\
\bottomrule
\end{tabular}
\end{table}

%======================================================================
\section{Results and Analysis}
\label{sec:results}
%======================================================================

\subsection{Main Results}
\label{sec:main_results}

We evaluate ASDA across two student models on both arithmetic and non-arithmetic tasks. For Haiku~3.5, we additionally include GEPA and ACE as training-free reference points; both achieve only marginal gains on FAMMA, which suggests the structural limitations of flat-text optimization.

\begin{table}[t]
\setlength{\tabcolsep}{4.5pt}
\renewcommand{\arraystretch}{1.15}
\caption{ASDA results across student models on FAMMA. For Haiku~3.5, GEPA and ACE are shown as reference baselines. All experiments use Sonnet~4.5 as the teacher model. WU = Warm-Up; E2 = best refinement epoch. $\Delta$ denotes absolute improvement in percentage points over each model's own baseline.}
\label{tab:main_results}
\centering
\begin{tabular}{ll cc cc}
\toprule
\multirow{2}{*}{\textbf{Student}} & \multirow{2}{*}{\textbf{Method}}
  & \multicolumn{2}{c}{\textbf{Arithmetic}}
  & \multicolumn{2}{c}{\textbf{Non-Arithmetic}} \\
\cmidrule(lr){3-4}\cmidrule(lr){5-6}
  & & \textbf{Acc.\ (\%)} & $\boldsymbol{\Delta}$
    & \textbf{Acc.\ (\%)} & $\boldsymbol{\Delta}$ \\
\midrule
\multirow{4}{*}{Haiku 3.5}
  & Baseline                         & 41.00 & ---   & 49.21 & --- \\
  & GEPA~\cite{agrawal2025gepa}      & 42.33 & +1.33 & 50.79 & +1.58 \\
  & ACE~\cite{ace2025}               & 44.30 & +3.30 & 49.60 & +0.39 \\
  & ASDA WU \textbf{(ours)}          & 49.67 & +8.67 & 51.98 & +2.78 \\
  & ASDA E2 \textbf{(ours)}          & \textbf{58.33} & \textbf{+17.33} & \textbf{55.16} & \textbf{+5.95} \\
\midrule
\multirow{3}{*}{Haiku 4.5}
  & Baseline                         & 64.67 & ---   & 57.14 & --- \\
  & ASDA WU \textbf{(ours)}          & 69.67 & +5.00 & 56.35 & $-$0.79 \\
  & ASDA E2 \textbf{(ours)}          & \textbf{70.66} & \textbf{+5.99} & \textbf{58.74} & \textbf{+1.60} \\
\bottomrule
\end{tabular}
\end{table}

ASDA consistently improves performance for Claude-family student models. On arithmetic, Haiku~3.5 gains +8.67\pp{} at warm-up and +17.33\pp{} after two refinement epochs. Haiku~4.5, despite its stronger 64.67\% baseline, achieves +5.99\pp{}---showing that ASDA adds value even for more capable models. Non-arithmetic gains are smaller but consistent for Haiku~3.5 (+2.78\pp{} warm-up, +5.95\pp{} at E2); Haiku~4.5 sees modest non-arithmetic improvement only after refinement (+1.60\pp{} at E2).

%Qwen Flash gains nothing on arithmetic (+0.00\pp{}) and regresses on non-arithmetic ($-$3.97\pp{}). 

%Its arithmetic errors are more evenly spread across failure types---concept confusion, unit errors, and code errors each account for roughly 30\%---compared to Haiku~3.5's concentrated failure profile (${\sim}$45\% concept confusion). When no single error type dominates, skills cannot target a large cluster of related failures, and the risk of introducing regressions on unrelated questions increases. This boundary condition suggests ASDA is most effective when the student exhibits systematic, clustered failure patterns rather than dispersed, heterogeneous errors.

\paragraph{Effect of iterative refinement.}
The warm-up phase targets the most frequent failure patterns and delivers the first large wave of gains. Iterative refinement then addresses residual failures that remain after initial skill injection. For arithmetic, accuracy improves from 49.67\% at warm-up to 54.67\% at epoch~1 and reaches 58.33\% at epoch~2. A smaller but consistent trend holds for non-arithmetic, where epoch~2 peaks at 55.16\%. Both domains regress at epoch~3 (arithmetic: 54.33\%; non-arithmetic: 51.98\%), indicating overfitting to residual training-set patterns after two refinement passes. In practice, two refinement epochs represent the optimal operating point.

\paragraph{Gains by question type.}
Skills consistently produce larger gains on multiple-choice questions than on open-ended ones. For Haiku~3.5 arithmetic at warm-up, MC accuracy rises by +14.39\pp{} compared to +3.73\pp{} for open-ended; Haiku~4.5 shows the same pattern (MC +7.91\pp{}, open-ended +2.48\pp{}). A skill that narrows the solution procedure is most useful when the answer space is already constrained to a few options. For open-ended generation, where the model must produce a free-form numerical or textual response, the guidance is less precise and the room for regression is higher.

\paragraph{Regressions.}

In sampled regression cases, the most common failure mode is skill-induced over-reasoning: the model follows the injected procedural guidance even when the question does not require it, elaborating beyond what is needed and revising an already-correct judgment. This occurs because the selector loads all skill files for a matched subfield as a bundle, so every question in that subfield receives guidance regardless of whether it needs it.

\subsection{Qualitative Analysis}

\begin{figure*}[ht!]
\centering
\includegraphics[width=0.95\textwidth]{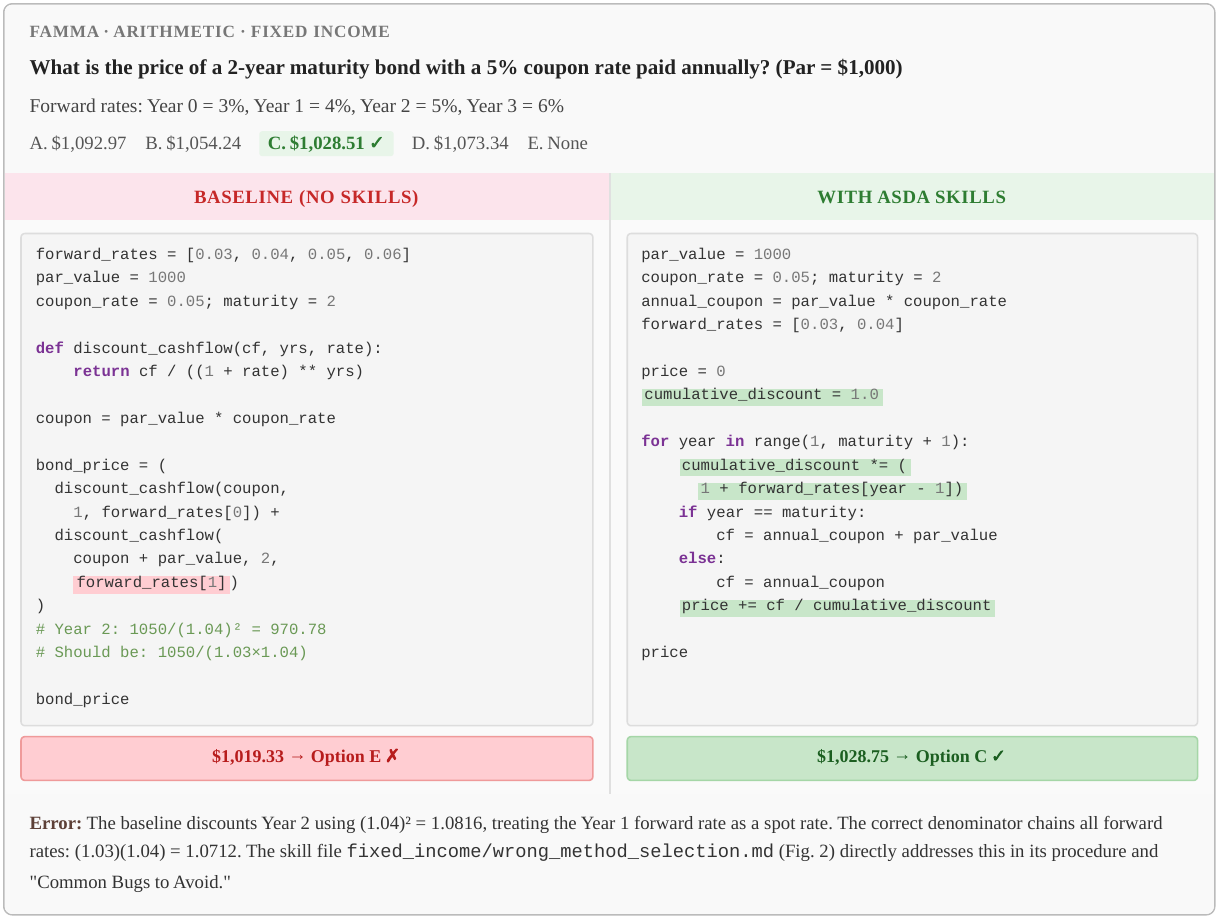}
\caption{Baseline vs.\ skill-augmented output on a FAMMA fixed 
income question (Haiku~3.5, warm-up). The skill file in 
Fig.~\ref{fig:skill-example} provides the domain-specific 
procedure that corrects the baseline error. The same skill file 
was credited with 7 additional fixes on related fixed income 
questions.}\label{fig:before-after}
\end{figure*}

To illustrate how skill artifacts operate at inference time, Figure~\ref{fig:before-after} presents an arithmetic case from the Haiku~3.5 warm-up evaluation where the baseline fails and skill injection succeeds. The same skill file was credited with 7 additional fixes on related fixed income questions, illustrating the reusability of subfield-specific skill artifacts.

%The question requires pricing a coupon bond using a schedule of forward rates --- a fixed income procedure that demands domain-specific knowledge about how forward rates compose into discount factors. Without skills, the model applies a generic discounting function that treats each forward rate independently, producing the wrong bond price. The skill file \texttt{fixed\_income/wrong\_method\_selection.md} (Fig.~\ref{fig:skill-example}), distilled from recurring failures on similar questions, encodes the correct sequential discounting procedure alongside a code template and a list of common errors. With this skill injected, the model follows the domain-specific procedure and arrives at the correct answer. The same skill file was the primary contributor to 7 other fixes on related fixed income questions, illustrating the reusability of subfield-specific skill artifacts.

\subsection{Where Does the Improvement Come From? Self-Teaching Ablation}
\label{sec:self_teaching}

ASDA's gains could come from two sources: knowledge contributed by a superior teacher model, or knowledge drawn from the training data itself. To separate these, we run a self-teaching configuration where the student model acts as its own teacher---it analyzes its own failures, builds the skill library, and then uses those skills at inference. No stronger model is involved.

\begin{table}[t]
\renewcommand{\arraystretch}{1.15}
\caption{Self-teaching ablation. Each model serves as both student and teacher; no superior model is involved. Full ASDA (Sonnet~4.5 teacher) results are shown for reference.}
\label{tab:self_teach}
\centering
\begin{tabular}{llccc}
\toprule
\textbf{Student} & \textbf{Domain} & \textbf{Baseline} & \textbf{With Skills} & $\boldsymbol{\Delta}$ \\
\midrule
\multicolumn{5}{l}{\textit{Full ASDA (Sonnet 4.5 teacher), for reference}} \\
Haiku 3.5 & Arithmetic & 41.00 & 49.67 & +8.67 \\
Haiku 3.5 & Non-Arith  & 49.21 & 51.98 & +2.78 \\
\midrule
\multicolumn{5}{l}{\textit{Self-Teaching (student = teacher)}} \\
Haiku 3.5  & Arithmetic & 41.00 & \textbf{47.33} & \textbf{+6.33} \\
Haiku 3.5  & Non-Arith  & 49.21 & 50.79          & +1.58 \\
%Sonnet 4.5 & Arithmetic & 73.33 & 70.33          & $-$3.00 \\
%Sonnet 4.5 & Non-Arith  & 69.05 & 67.86          & $-$1.19 \\
%Sonnet 4.6 & Arithmetic & 74.67 & 69.00          & $-$5.67 \\
%Sonnet 4.6 & Non-Arith  & 64.68 & \textbf{70.24} & \textbf{+5.56} \\
\bottomrule
\end{tabular}
\end{table}

For Haiku~3.5 arithmetic, self-teaching achieves +6.33\pp{}---73\% of the +8.67\pp{} gain from using a Sonnet~4.5 teacher. This shows that most of the benefit comes from the training data, not from the teacher's superior knowledge. The training questions provide only the question text and the correct answer---no worked solutions or step-by-step reasoning---yet seeing where it consistently goes wrong is enough for the model to identify its recurring failure patterns and synthesize skills to address them. The model already possesses much of the relevant domain knowledge; the distillation process gives it the structure to apply that knowledge reliably. The remaining 2.34\pp{} gap reflects the teacher's contribution---a stronger model produces sharper failure diagnoses and more precise skill formulations. The same pattern holds for non-arithmetic (+1.58\pp{} self-teaching vs.\ +2.78\pp{} with a Sonnet teacher).

%\paragraph{Self-teaching in stronger models.}
%Sonnet~4.5 and Sonnet~4.6 both regress on arithmetic ($-$3.00\pp{} and $-$5.67\pp{}). At baselines above 73\%, the model already answers most arithmetic questions correctly; the remaining failures are sparse and varied, and skills synthesized from this thin signal tend to over-constrain the model on questions it previously handled well. Sonnet~4.6 non-arithmetic is an exception, achieving +5.56\pp{}---the best non-arithmetic result in this study. Here the 64.68\% baseline leaves more room for improvement, and the residual failures are clustered enough for distillation to produce useful skills. The contrast with Sonnet~4.5 ($-$1.19\pp{} non-arithmetic at a 69.05\% baseline) suggests that baseline accuracy alone does not fully predict self-teaching success; the structure and consistency of the residual failures also matter.

\subsection{Are Skills Student-Specific? Cross-Transfer Experiment}
\label{sec:cross_transfer}

The self-teaching results raise a natural follow-on question: once skills are distilled for one model, can they benefit another? We apply skills generated from Haiku~3.5's failures to Haiku~4.5 (a stronger model in the same family) and compare against skills derived from Haiku~4.5's own failures.

\begin{table}[t]
\renewcommand{\arraystretch}{1.15}
\caption{Skill portability: Haiku~4.5 arithmetic (300 eval questions). Own skills are generated from Haiku~4.5's failures; cross-transfer uses skills generated from Haiku~3.5's failures.}
\label{tab:cross_transfer}
\centering
\begin{tabular}{lccccc}
\toprule
\textbf{Skills Source} & \textbf{Baseline} & \textbf{With Skills} & $\boldsymbol{\Delta}$ & \textbf{MC} $\boldsymbol{\Delta}$ & \textbf{Open} $\boldsymbol{\Delta}$ \\
\midrule
Own skills (H4.5)            & 64.67 & \textbf{69.67} & \textbf{+5.00} & +7.91 & +2.48 \\
H3.5 skills (cross-transfer) & 64.67 & 62.33          & $-$2.33        & +2.16 & $-$6.21 \\
\bottomrule
\end{tabular}
\end{table}

Cross-model transfer produces a net regression of $-$2.33\pp{}, driven by a sharp open-ended decline ($-$6.21\pp{}) that overwhelms modest MC gains (+2.16\pp{}). In contrast, Haiku~4.5 with its own ASDA skills gains a consistent +5.00\pp{} across both question types.

The practical implication is direct: skills should be generated for each
deployed model independently. Reusing skills across model generations is
unlikely to yield positive returns and can actively harm the stronger model.
As shown in Table~\ref{tab:cost}, student-specific warm-up distillation
costs under \$12, with two refinement epochs adding \$89--135, making
per-model regeneration operationally feasible at deployment scale.

%======================================================================
\section{Discussion and Conclusion}
\label{sec:discussion}
%======================================================================
 
\paragraph{Discussion.}
Two experimental findings reveal the underlying mechanism of ASDA.
The self-teaching result---where a model acting as its own teacher recovers 73\% of the full arithmetic gain---shows that the primary source of improvement is not the teacher's superior knowledge but the structure imposed by the distillation process itself.
Enumerating failure patterns across a training set forces the model to externalize domain knowledge it already implicitly holds but cannot reliably apply during single-pass inference.
The cross-transfer failure reinforces this from the opposite direction: skills are not generic domain knowledge that transfers across models, but artifacts of a specific model's failure distribution.
Applying one model's skills to a stronger model constrains it by the weaker model's blind spots.
Together, these results characterize skills as \emph{model-specific failure remedies}---a distinction with direct consequences for how skill libraries should be managed in practice.

For organizations that rely on commercial LLMs through black-box APIs, ASDA offers a concrete operational path, and credit analysis is a representative case. Drafting a credit assessment requires spreading financial statements, computing coverage and leverage ratios, testing covenant headroom, and applying the institution's own credit policy and regulatory criteria before arriving at a rating. This reasoning is procedural and quantitative, backed by a labeled history of completed assessments to distill from, and the recurring errors---an incorrect ratio definition, a misapplied policy threshold, an overlooked covenant---cluster cleanly by type. Distilling skills from that history yields a library that encodes the correct procedure for each failure mode---one that credit-risk officers can review, compliance can certify against policy, and engineering can version-control and regenerate when the base model is upgraded, without re-validating opaque fine-tuned weights. The approach suits workflows whose expertise is codified in policy and procedure more than those resting on tacit, case-specific judgment such as fraud investigation. For any regulated workflow with similar structure, the skill library provides an adaptation path that remains fully auditable---an assurance that fine-tuning cannot offer.

ASDA's gains are largest when failure patterns are clustered and the task has well-defined procedural structure, as with arithmetic reasoning via Program-of-Thought.
When errors are more dispersed---as in non-arithmetic tasks---the signal available for distillation is weaker and regression risk increases.
This boundary condition suggests that diagnostic quality (how cleanly errors cluster by type and subfield) predicts adaptation success alongside raw model capability.
 
\paragraph{Limitations.}
All experiments use a single benchmark (FAMMA) and a single model
family (Anthropic's Claude). While the consistency of gains across
two student models and two task types provides some evidence of
robustness, it does not establish whether the error taxonomy, skill
format, or refinement dynamics transfer to other domains or model
architectures. FAMMA's reliance on OCR-extracted text introduces an
additional confound: skills distilled from noisy transcriptions may
encode artifact-specific heuristics, meaning the regression rates we
report may overestimate what would be observed on cleaner datasets.
More broadly, all evaluation is benchmark-based; whether the observed
accuracy gains translate to practical value in production financial
workflows remains an open question.
 
\paragraph{Future Work.}
\emph{Cross-domain transfer.} Evaluating on healthcare and legal
reasoning benchmarks would address the single-benchmark limitation
and test whether skills transfer across domains. These domains are
procedural, knowledge-intensive, and carry auditability requirements
that align with the skill library's inspectable format.

\emph{Cross-model generalization.} Our results are limited to
Anthropic's Claude family in both student and teacher roles.
Evaluating with models from other families (e.g., GPT, DeepSeek)
would clarify whether the framework's gains depend on
family-specific instruction-following behavior or generalize more
broadly.

\emph{Finer-grained skill selection.} The current pipeline generates
10--30 skill files per configuration, and the selector loads all files
for a matched subfield as a bundle. A finer-grained approach that
selects at the pattern level, combined with pruning of low-value
skills, could reduce the regressions caused by unnecessary procedural
guidance while sharpening routing precision.

Preliminary experiments on cross-domain and cross-model evaluation
are underway.

\paragraph{Conclusion.}
We have shown that failure-driven distillation can externalize latent domain knowledge into an explicit, inspectable form that single-pass inference cannot access---without modifying model weights. Our ablations further reveal that the resulting skills function as \emph{model-specific failure remedies}: artifacts of a particular model's failure distribution that cannot be transferred across model generations, but can be cheaply regenerated for any deployed model given only a labeled domain dataset. Warm-up distillation costs under \$12, with optional iterative refinement adding \$89--135 (Table~\ref{tab:cost}). The skill library is not a better prompt: it is a new representational layer between the raw model and its deployment context, one that can be version-controlled, audited, and regenerated for successor models.
Whether this layer generalizes to other knowledge-intensive domains, and how far self-teaching can substitute for stronger supervision, remain the central open questions.
\begin{credits}

\subsubsection{Generative AI Disclosure}
The authors used generative AI tools to assist with manuscript 
editing and experimental pipeline development. All scientific content---hypotheses, experimental designs, and 
conclusions---is entirely the work of the human authors.
\subsubsection{\discintname}
The authors have no competing interests to declare that are relevant to the content of this article.
\end{credits}

%----------------------------------------------------------------------
% REFERENCES
%----------------------------------------------------------------------
\bibliographystyle{splncs04}

\end{document}